\documentclass[runningheads]{llncs}
\usepackage{graphicx}

\usepackage[export]{adjustbox}
\usepackage{graphicx}
\usepackage{url}
\usepackage{float}
 \usepackage{listings, paralist}
 \PassOptionsToPackage{hyphens}{url}
 \usepackage{hyperref}

\usepackage[T1]{fontenc}
\usepackage[scaled=0.85]{beramono}
\usepackage{listings}
\usepackage{tabularx}
\usepackage{comment}
\usepackage{todonotes}
\usepackage{wrapfig}

\definecolor{lightBlue}{RGB}{0, 153, 255}
\definecolor{lightRed}{RGB}{204, 0, 255}

\lstdefinelanguage{Manchester}
{
    sensitive = true,
    keywords = [1]{Class, EquivalentTo, SubClassOf},
    morekeywords = [2]{and, or},
    morekeywords = [3]{some, value},
    keywordstyle=[2]\textbf,
    keywordstyle=[2]\color{lightBlue}\textbf,
    keywordstyle=[3]\color{lightRed}\textbf,
    morestring=[b]''
}

\begin{document}
\title{Explanation Ontology: A Model of Explanations for User-Centered AI}
%
%
\author{Shruthi Chari\inst{1}\orcidID{0000-0003-2946-7870} \and
Oshani Seneviratne\inst{1}\orcidID{0000-0001-8518-917X} \and
Daniel M. Gruen\inst{2}\orcidID{0000-0003-0155-3777} \and Morgan A. Foreman\inst{2}\orcidID{0000-0002-2739-5853} \and Amar K. Das\inst{2}\orcidID{0000-0003-3556-0844} \and Deborah L. McGuinness\inst{1}\orcidID{0000-0001-7037-4567} }
\authorrunning{S. Chari et al.}

\institute{
Rensselaer Polytechnic Institute, Troy, NY, USA \\
\email{\{charis, senevo\}@rpi.edu,dlm@cs.rpi.edu}\\
\and
IBM Research, Cambrdige, MA, USA \\
\email{\{daniel\_gruen, morgan.foreman, amardas\}@us.ibm.com}}
\maketitle              
\begin{abstract}
Explainability has been a goal for Artificial Intelligence (AI) systems since their conception, with the need for explainability growing as more complex AI models are increasingly used in critical, high-stakes settings such as healthcare. Explanations have often added to an AI system in a non-principled, post-hoc manner. With greater adoption of these systems and emphasis on user-centric explainability, there is a need for a structured representation that treats explainability as a primary consideration, mapping end user needs to specific explanation types and the system's AI capabilities. We design an explanation ontology to model both the role of explanations, accounting for the system and user attributes in the process, and the range of different literature-derived explanation types.  We indicate how the ontology can support user requirements for explanations in the domain of healthcare. We evaluate our ontology with a set of competency questions geared towards a system designer who might use our ontology to decide which explanation types to include, given a combination of users' needs and a system's capabilities, both in system design settings and in real-time operations. Through the use of this ontology, system designers will be able to make informed choices on which explanations AI systems can and should provide. 

\textbf{Resource:} \url{https://tetherless-world.github.io/explanation-ontology}

\keywords{Explainable AI \and Explanation Ontology \and Modeling of Explanations and Explanation Types \and Supporting Explainable AI in Clinical Decision Making and Decision Support}
\end{abstract}
\section{Introduction} \label{sec:introduction}
Explainability has been a key focus area of Artificial Intelligence (AI) research, from expert systems, cognitive assistants, the Semantic Web, and more recently, in the machine learning (ML) domain. In our recent work \cite{chari2020foundationschapter},
we show that advances in explainability have been coupled with advancements in the sub-fields of AI. For example, explanations in second-generation expert systems typically address \textit{What, Why, and How} questions \cite{swartout1993explanation,dhaliwal1996use}. With ML methods, explainability has focused on interpreting the functioning of black-box models, such as identifying the input features that are associated the most with different outputs \cite{lou2012intelligible,lipton2016mythos}. However, while explanations of the ``simplified approximations of complex decision-making functions'' \cite{mittelstadt2019explaining} are important, they do not account for ``specific context and background knowledge'' \cite{paez2019pragmatic} that users might possess, and hence, are often better suited for experts or debugging purposes. Several researchers have written about this shortcoming \cite{mittelstadt2019explaining,lipton2016mythos}, and the fallacy of associating explainability to be solely about model transparency and interpretability \cite{holzinger2017we}. Given this shift in focus of explainable AI, due to the adoption of AI in critical and user-facing fields such as healthcare and finance, researchers are drawing from adjacent ``explanation science'' fields to make explainable AI more usable \cite{miller2019explanation}. The term ``explanation sciences'' was introduced by Mittlestadt et al. to collectively refer to the fields of ``law, cognitive science, philosophy, and the social sciences.'' \cite{mittelstadt2019explaining}

Recent review papers \cite{mittelstadt2019explaining,biran2017explanation} point out that explainability is diverse, serving and addressing different purposes, with user-specific questions and goals. Doshi et al. \cite{doshi2017towards} propose a set of questions beyond the \textit{What, Why, How} questions that need to be addressed by explanations: ``What were the main factors in a decision?'', ``Would changing a certain factor have changed the decision?'' and ``Why did two similar-looking cases get different decisions or vice versa?'' Other researchers, like Wang et al. \cite{wang2019designing}, in their conceptual framework linking human reasoning methods to explanations generated by systems, support various explainability features, such as different ``intelligibility queries.'' Lim and Dey observed these intelligibility queries during their user study where they were studying for mechanisms to improve the system's intelligibility (comprehensibility) by looking to gain user's trust and seeking to avoid situations that could lead to a ``mismatch between user expectation and system behavior'' \cite{lim2009and}. Support for such targeted provisions for explanations would enable the creation of ``explainable knowledge-enabled systems", a class of systems we defined in prior work \cite{chari2020foundationschapter} as, ``AI systems that include a representation of the domain knowledge in the field of application, have mechanisms to incorporate the \textit{users' context}, are \textit{interpretable}, and host \textit{explanation facilities} that generate \textit{user-comprehensible, context-aware, and provenance-enabled} explanations of the mechanistic functioning of the AI system and the knowledge used.''

Currently, a single class of AI models, with their specific focus on particular problems that tap into specific knowledge sources, cannot wholly address these broad and diverse questions. The ability to address a range of user questions points to the need for providing explanations as a service via a framework that interacts with multiple AI models with varied strengths. To achieve this flexibility in addressing a wide range of user questions, we see a gap in semantic support for the generation of explanations that would allow for explanations to be a core component of AI systems to meet users' requirements. We believe an ontology, a machine-readable implementation, can help system designers, and eventually a service, to identify and include methods to generate a variety of explanations that suit users' requirements and questions. While there have been a few efforts to establish connections between explanation types and the mechanisms that are capable of generating them \cite{wang2019designing,arrieta2020explainable}, these efforts are either not made available in machine-readable formats or not represented semantically. The lack of a semantic representation that would offer support for modeling explanations makes it difficult for a system designer to utilize their gathered user requirements as a means to build in explanation facilities into their systems. 

We first present related work on taxonomies of explanation (Section \ref{sec:related_work}). We then introduce the design of an explanation ontology (Section \ref{sec:ontology}) that treats explanations as a primary consideration of AI system design and that can assist system designers to  capture and structure the various components necessary to enable the generation of user-centric explanations computationally. These components and the attributes of explanations are gathered from our literature review of AI \cite{chari2020foundationschapter,chari2020directionschapter}, associated ``explanation sciences'' domains
including social sciences and philosophy. We use the results of a previously conducted user-centered design study, which used a prototype decision-support system for diabetes management, to demonstrate the usage of our ontology for the design and implementation of an AI system (Section \ref{sec:instantiations}). Finally, we evaluate our ontology's competency in assisting system designers by our ontology's ability to support answering a set of questions aimed at helping designers build
``explainable, knowledge-enabled,'' AI systems (Section \ref{sec:evaluation}).
\section{Related Work} \label{sec:related_work}
While there have been several taxonomies proposed within explainable AI, we limit our review to taxonomies that are closest to our focus, in that they catalog AI models and the different types of explanations they achieve \cite{arya2019one,arrieta2020explainable}, or ones that capture user-centric aspects of explainability \cite{wang2019designing}. Recently, Arya et al. \cite{arya2019one} developed a taxonomy for explainability, organizing ML methods and techniques that generate different levels of explanations, including \textit{post-hoc} (contain explanations about the results or model functioning), \textit{local} (about a single prediction), and \textit{general} (describes behavior of entire model) explanations, across various modalities (interactive/visual, etc.). Their taxonomy has been used as a base for the AI Explainability 360 (AIX 360) toolkit \cite{aix360} to recommend applicable, explainable AI methods to users. Their taxonomy only considers attributes of explanations from an implementation and data perspective and does not account for end-user-specific requirements. Further, their taxonomy, implemented as a decision tree, lacks a semantic mapping of the terms involved, which makes it hard for system designers to extend this taxonomy flexibly or to understand the interaction between the various entities involved in the generation of explanations. In our ontology, we provide a semantic representation that would help system designers support and include different explanation types in their system, while accounting for both system and user attributes.

Similarly, Arrieta et al. \cite{arrieta2020explainable} have produced a taxonomy, mapping ML models (primarily deep learning models) to the explanations they produced and the features within these models that are responsible for generating these explanations. Their taxonomy covers different types of explanations that are produced by ML models, including \textit{simplification, explanation by examples, local explanations, text explanations, visual explanations, feature relevance, and explanations by transparent models}. However, in their structural taxonomy, due to the lack of a semantic representation, they often refer to explanation types and the modalities in which they are presented interchangeably. In addition, the explanations they cover are tightly coupled with the capabilities of ML models and they do not explore other aspects that could be included in explanations, such as different forms of knowledge, to make them amenable to end users. Through our ontology, we address this gap by incorporating diverse aspects of explanaibility that are relevant to supporting the generation of user-centric explanation types (e.g., counterfactual, contrastive, scientific, trace based explanations, etc.) that address different user goals beyond model interpretability. 

Wang et al. have developed a ``Conceptual Framework for Reasoned Explanations'' to describe ``how human reasoning processes inform'' explainable AI techniques \cite{wang2019designing}. Besides supporting connections between how humans reason and how systems generate an explanation, this conceptual framework also draws parallels between various aspects of explainability, such as between explanation goals and explanation types, human reasoning mechanisms and AI methods, explanation types and intelligibility queries. While they cover some explanation types and point to the need for ``integrating multiple explanations,'' we support a broader set of literature-derived explanation types via our ontology. Also, it remains unclear as to whether their framework is in a machine-readable format that can be used to support system development. Within our ontology, we model some of these explainability aspects that are captured in their framework, including explanation types, explanation goals, and different modalities.

Tiddi et al. \cite{tiddi2015ontology} created an ontology design pattern (ODP) to represent explanations, and showcased the ability of the ODP to represent explanations across ``explanation sciences,'' such as linguistics, neuroscience, computer science, and sociology. In this ODP, they associate explanations with attributes, including the situation, agents, theory, and events. Additionally, they provide support for the association of explanations with \textit{explanandum} (the fact or event to be explained) and \textit{explanan} (that which does the explaining) to associate explanations with some premise to the \textit{explanandum}. Their contribution is a general-purpose ODP, however, it cannot be applied as is in practice, due to the difficulty in condensing explanations to their suggested form of \textit{ <explanans (A), posterior explanandum (P), theory (T), and situational context (C)>}, without the background understanding of how these entities were generated in their field of application. In our ontology, we reuse classes and properties from this ODP, where applicable, and expand on their mappings to support the modeling of explanations generated via computational processes and that address the users' questions, situations, and contexts.  
\section{Explanation Ontology} \label{sec:ontology} 
As we have discussed in Section \ref{sec:introduction} and \ref{sec:related_work}, explainability serves different purposes. Given this diversity, there is a need for a semantic representation of explanations that models associations between entities and attributes directly and indirectly related to explanations from the system as well as user standpoints. In designing our ``Explanation Ontology,'' (EO) we have used both bottom-up and top-down modeling approaches. We undertook a bottom-up literature review to primarily identify different explanation types and their definitions in the literature. We utilize our literature review as a base for our modeling and  use a top-down approach to refine the modeling by analyzing the usage of different explanation types by clinicians during a requirements gathering session we conducted. In Section \ref{subsec:modeling}, we describe our modeling, then, in Section \ref{subsec:explanationtypes}, we showcase our representation of literature-derived explanation types using this modeling.

\subsection{Ontology Composition} \label{subsec:modeling}
\begin{figure}[!hbt]
\includegraphics[width=1.0\linewidth]{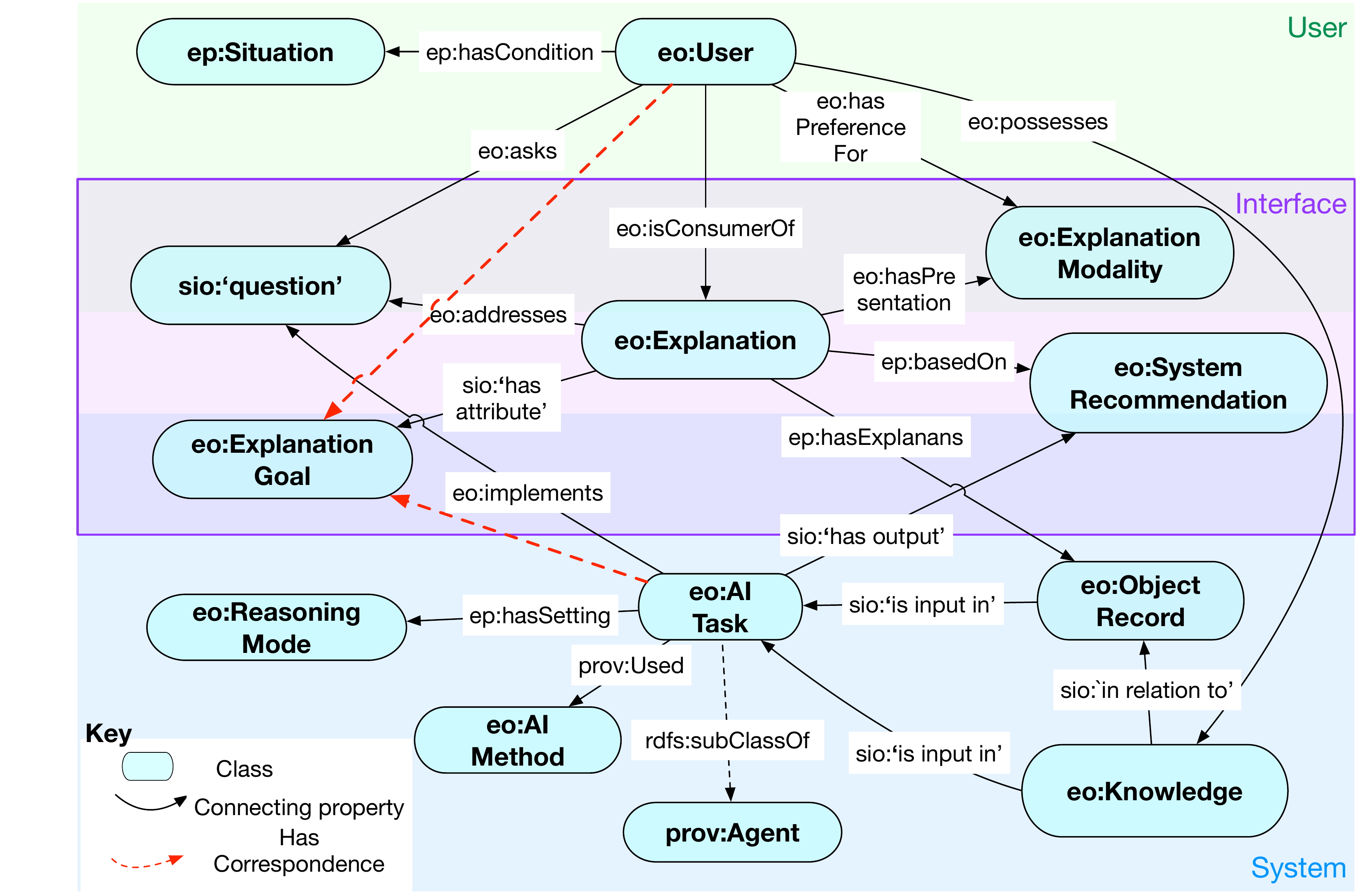}
\caption{A conceptual overview of our explanation ontology, capturing attributes of explanations to allow them to be assembled by an AI Task, used in a system interacting with a user. We depict user-attributes of explanations in the upper portion (green highlight), system-attributes in the lower portion (blue highlight), and attributes that would be visible in a user interface are depicted in the middle portion in purple. } \label{fig:conceptmap} 
\end{figure}

We design our ontology around the central explanation class (ep:Explanation) and include entities and attributes that we see occurring often in the literature as explanation components. 
In Fig. \ref{fig:conceptmap}, we present a conceptual overview of our ontology and depict associations necessary to understand its coverage. In Table \ref{tab:prefixtable}, we list ontology prefixes that we use to refer to classes and properties.

\begin{table}[hbt!]
\centering
\caption{List of ontology prefixes used in the paper.}
\label{tab:prefixtable}
\resizebox{\textwidth}{!}{%
\begin{tabular}{|l|l|l|}
\hline
\begin{tabular}[c]{@{}l@{}}Ontology \\ Prefix\end{tabular} & Ontology             & URI                           \\ \hline
sio & \begin{tabular}[c]{@{}l@{}}SemanticScience Integrated\\ Ontology\end{tabular} & \url{http://semanticscience.org/resource/}              \\ \hline
prov                                                       & Provenance Ontology  & \url{http://www.w3.org/ns/prov-o\#} \\ \hline
eo                                                         & Explanation Ontology & \url{https://purl.org/heals/eo\#}   \\ \hline
ep  & \begin{tabular}[c]{@{}l@{}}Explanation Patterns \\ Ontology\end{tabular}      & \url{http://linkedu.eu/dedalo/explanationPattern.owl\#} \\ \hline
\end{tabular}%
}
\end{table}

In our ontology, we build on the class and property hierarchies provided by Tiddi et al.'s explanation ODP \cite{tiddi2015ontology} and the general-purpose SemanticScience Integrated Ontology (SIO) \cite{dumontier2014semanticscience}. When referencing classes and properties from the SemanticScience Integrated Ontology (SIO) that use numeric identifiers, we follow the convention used in their paper \cite{dumontier2014semanticscience} by referring to classes and properties via their labels. E.g., sio:`in relation to'. 

We introduce classes and properties as necessary to construct a model of explanations that supports property associations between explanations (ep:Explanation), the AI Task (eo:AITask) that generated the recommendation (eo:SystemRecommendation) upon which the explanation is based (ep:isBasedOn) and the end-user (eo:User) who consumes them. In our modeling, we note that explanations are dependent on (ep:isBasedOn) both system recommendations as well as implicit/explicit knowledge (eo:knowledge) available to systems or possessed by users. In addition, we also model that the knowledge available to the system can be in relation to (sio:`in relation to') various entities, such as the domain knowledge, situational knowledge of the end-user (eo:User), and knowledge about a record of the central object (eo:ObjectRecord) fed into the system (e.g., as we will see in Section \ref{sec:instantiations}, a patient is a central object in a clinical decision support system). We also model that explanations address a question (sio:`question') posed by end-users, and these questions are implemented (eo:implements) by AI Tasks that generate recommendations. 

AI Tasks (eo:AITask) can be thought of as analogous with different reasoning types (i.e., inductive, deductive, abductive or a hybrid reasoning strategy) and that are implemented by different AI methods (eo:AIMethod) (e.g., similarity algorithms, expert systems) to arrive at recommendations. This decomposition of an AI Task to methods is inspired by Tu et al.'s research in the problem-solving domain \cite{tu1995ontology} of creating domain-independent AI systems that can be instantiated for specific domains. Besides capturing the interplay between an eo:AITask and its implementation, an eo:AIMethod, we also model that an AI Task is implemented in a particular reasoning mode (eo:ReasoningMode) of the system which dictates the overall execution strategy. We believe our approach to supporting the different granularities of work separation within an AI system can be valuable to building AI systems with hybrid reasoning mechanisms capable of generating different explanation types. In addition to capturing the situational context of the user, we also support modeling their existing knowledge and preferences for different forms for presentations of explanation (eo:ExplanationModality).

In the rest of this paper, we refer to ontology classes within single quotes and italicize them (e.g., \textit{`knowledge'}) to give readers an idea of the coverage of our ontology. While, we only depict the top-level classes associated with explanations in Fig. \ref{fig:conceptmap}, through representations of different explanation types and an example from the clinical requirements gathering session, we show how system designers can associate explanations with more specific subclasses of entities, such as with particular forms of \textit{`knowledge'}, \textit{`AI Task'}s, and \textit{`AI methods'}. Further, we maintain definitions and attributions for our classes via the usage of terminology from the DublinCore \cite{dct} and Prov-O \cite{lebo2013prov} ontologies. 

\subsection{Modeling of Literature-derived Explanation Types}
\label{subsec:explanationtypes}
We previously created a taxonomy of literature-derived, explanation types \cite{chari2020directionschapter} with refined definitions of the nine explanation types.
We leverage the mappings provided within EO, and knowledge of explanation types from our taxonomy, to represent each of these explanation types as subclasses of the explanation class (ep:Explanation). These explanation types serve different user-centric purposes, are differently suited for users'
\textit{`situations,'} context and \textit{`knowledge,'} are generated by various \textit{`AI Task'} and \textit{`methods,'} and have different informational needs. Utilizing the classes and properties supported within our ontology, we represent the varied needs for the generation of each explanation type, or the sufficiency conditions for each explanation type, as OWL restrictions. 

In Table \ref{tab:explanationtypes}, we present an overview of the different explanation types along with their descriptions and sufficiency conditions. In Listing \ref{lst:contextualexp}, we present an RDF representation of a \textit{`contextual explanation,'} depicting the encoding of sufficiency conditions on this class. 
\lstset{language=Manchester, basicstyle=\ttfamily\fontsize{9}{10}\selectfont, columns=fullflexible, xleftmargin=5mm, framexleftmargin=5mm, numbers=left, stepnumber=1, breaklines=true, breakatwhitespace=false, numberstyle=\ttfamily\fontsize{9}{10}\selectfont, numbersep=5pt, tabsize=2, frame=lines, captionpos=t, caption={OWL expression of the representation of a \textit{`contextual explanation'} (whose sufficiency conditions can be referred to from Table \ref{tab:explanationtypes}) in Manchester syntax. In this snippet, we show the syntax necessary to understand the composition of the \textit{`contextual explanation'} class in reference to the classes and properties introduced in Fig. \ref{fig:conceptmap}.
}, label=lst:contextualexp}
\begin{lstlisting} 
Class: eo:ContextualExplanation
  EquivalentTo:
    (isBasedOn some eo:`System Recommendation')
    and (
    (ep:isBasedOn some 
        (eo:'Contextual Knowledge'
         and (sio:`in relation to' some ep:Situation))) or 
    (ep:isBasedOn some (`Contextual Knowledge'
         and (sio:`in relation to' some eo:`Object Record'))))
  SubClassOf:
    ep:Explanation

Class: ep:Explanation
    SubClassOf: 
        sio:`computational entity',
        ep:isBasedOn some eo:Knowledge,
        ep:isBasedOn some eo:SystemRecommendation,
        ep:isConceptualizedBy some eo:AITask
\end{lstlisting}

\begin{table}[hbt!]
\centering
\caption{An overview of explanation types against simplified descriptions of their literature-synthesized definitions, and natural language descriptions of sufficiency conditions. Within the explanation type description we also include a general prototypical question that can be addressed by each explanation type. Further, within the sufficiency conditions, we highlight ontology classes using single quotes and italics.}
\label{tab:explanationtypes}
\resizebox{\textwidth}{!}{%
\begin{tabular}{|l|l|l|}
\hline
\multicolumn{1}{|c|}{\textbf{\large{Explanation Type}}} &
  \multicolumn{1}{c|}{\textbf{\large{Description}}} &
  \multicolumn{1}{c|}{\textbf{\large{Sufficiency Conditions}}} \\ \hline
\textbf{Case Based} &
  \begin{tabular}[c]{@{}l@{}}Provides solutions that are based on actual prior cases \\ that can be presented to the user to provide compelling \\ support for the system’s conclusions, and may involve\\ analogical reasoning, relying on similarities between\\ features of the case and of the current situation.\\
  \textbf{``To what other situations has this }\\ \textbf{recommendation been applied?''}\end{tabular} &
  \begin{tabular}[c]{@{}l@{}}Is there at least one other prior case (\textit{`object record'})  \\ similar to this situation that had an \textit{`explanation'}?\\ Is there a similarity between this case, and \\ that other case?\end{tabular} \\ \hline
\textbf{Contextual} &
  \begin{tabular}[c]{@{}l@{}}Refers to information about items other than the \\ explicit inputs and output, such as information about \\ the user, situation, and broader environment that \\ affected the computation. \\
  \textbf{``What broader information about the current} \\ \textbf{situation prompted the suggestion of this} \\ \textbf{recommendation?''} \end{tabular} &
  \begin{tabular}[c]{@{}l@{}}Are there any other extra inputs that are not \\ contained in the \textit{`situation'} description itself?\\ And by including those, can better insights \\ be included in the \textit{`explanation'}?\end{tabular} \\ \hline
\textbf{Contrastive} &
  \begin{tabular}[c]{@{}l@{}}Answers the question “Why this output instead of that \\ output,” making a contrast between the given output \\ and the facts that led to it (inputs and other \\ considerations),  and an alternate output of interest and \\ the foil (facts that would have led to it). \\ \textbf{``Why choose option A} \\\ \textbf{over option B that I typically choose?''}\end{tabular} &
  \begin{tabular}[c]{@{}l@{}}Is there a \textit{`system recommendation'} that was made \\ (let’s call it A)? What facts led to it?\\ Is there another \textit{`system recommendation'} that \\ could have happened or did occur, (let’s call it B)? \\ What was the \textit{`foil'} that led to B?\\ Can A and B be compared?\end{tabular} \\ \hline
\textbf{Counterfactual} &
  \begin{tabular}[c]{@{}l@{}}Addresses the question of what solutions would have \\ been obtained with a different set of inputs than\\ those used. \\ 
  \textbf{``What if input A was over 1000?”}
  \end{tabular} &
  \begin{tabular}[c]{@{}l@{}}Is there a different set of inputs that can be \\ considered?\\ If so what is the alternate \textit{`system recommendation'}?\end{tabular} \\ \hline
\textbf{Everyday} &
  \begin{tabular}[c]{@{}l@{}}Uses accounts of the real world that appeal to the user, \\ given their general understanding and knowledge. \\ 
  \textbf{``Why does option A make sense”}\end{tabular} &
  \begin{tabular}[c]{@{}l@{}}Can accounts of the real world be \\ simplified to appeal to the user based on\\ their general understanding and \textit{`knowledge'}?\end{tabular} \\ \hline
\textbf{Scientific} &
  \begin{tabular}[c]{@{}l@{}}References the results of rigorous scientific methods, \\ observations, and measurements. \\
  \textbf{``What studies have backed this} \\  \textbf{recommendation?''} \end{tabular} &
  \begin{tabular}[c]{@{}l@{}}Are there results of rigorous \textit{`scientific} \\ \textit{methods'} to explain the situation?\\ Is there \textit{`evidence'} from the literature to\\ explain this \textit{`situation'}?\end{tabular} \\ \hline
\textbf{Simulation Based} &
  \begin{tabular}[c]{@{}l@{}}Uses an imagined or implemented imitation of a \\ system or process and the results that emerge from \\ similar inputs. 
  \\ \textbf{``What would happen if this recommendation} \\ \textbf{is followed?''} \end{tabular} &
  \begin{tabular}[c]{@{}l@{}}Is there an \textit{`implemented'} \\ imitation of the \textit{`situation'} at hand?\\ Does that other scenario have inputs similar\\ to the current \textit{`situation'}?\end{tabular} \\ \hline
\textbf{Statistical} &
  \begin{tabular}[c]{@{}l@{}}Presents an account of the outcome based on data about\\ the occurrence of events under specified \\ (e.g., experimental) conditions. Statistical explanations \\ refer to numerical evidence on the likelihood of factors \\ or processes influencing the result. \\
  \textbf{``What percentage of people with } \\
  \textbf{this condition have recovered?''} \end{tabular} &
  \begin{tabular}[c]{@{}l@{}}Is there \textit{`numerical evidence'}/likelihood \\ account of the \textit{`system recommendation'} based on \\ data about the occurrence of the outcome \\ described in the recommendation?\end{tabular} \\ \hline
\textbf{Trace Based} &
  \begin{tabular}[c]{@{}l@{}}Provides the underlying sequence of steps used by the \\ system to arrive at a specific result, containing the line \\ of reasoning per case and addressing the question of \\ why and how the application did something. \\ \textbf{``What steps were taken by the system to} \\ \textbf{generate this recommendation?''} \end{tabular} &
  \begin{tabular}[c]{@{}l@{}}Is there a record of the underlying sequence\\ of steps (\textit{`system trace'}) used by the \textit{`system'} to \\ arrive at a specific \textit{`recommendation'}?\end{tabular} \\ \hline
\end{tabular}%
}
\end{table}

\section{Clinical Use Case} \label{sec:instantiations}
We demonstrate the use of EO in the design and operations of an AI system to support treatment decisions in the care of patients with diabetes. We previously conducted a two-part user-centered design study that focused on determining which explanations types are needed within such a system.  In the first part of the study, we held an expert panel session with three diabetes specialists to understand their decision-support needs when applying guideline-based recommendations in diabetes care. We then used the requirements gathered from this session to design a prototype AI system. In the second part, we performed  cognitive walk-throughs of the prototype to understand what reasoning strategies clinicians used and which explanations were needed when presented with a complex patient. 

In modeling the reasoning strategies that need to be incorporated into this system design, we found that the Select-Test (ST) model by Stefanoli and Ramoni \cite{stefanelli1992epistemological} mirrored the clinician's approach.  Applying their ST model, we can organize the clinical reasoning strategy within the system design based on types of \textit{`reasoning mode,'} such as differential diagnosis or treatment planning.  Each of these modes can be associated with AI tasks, such as \textit{ranking} that creates a preferential order of options like diagnoses or treatments, \textit{deduction} that predicts consequences from hypotheses, \textit{abstraction} that identifies relevant clinical findings from observations, and \textit{induction} that selects the best solution by matching observations to the options or requests new information where necessary. Each of these AI tasks can generate system recommendations that requires explanations from the clinicians.

We discovered that, of the types of explanations listed in Table 1, everyday and contextual explanations were required more than half the time. We noted that clinicians were using a special form of everyday explanations, specifically their experiential knowledge or \textit{`clinical pearls'} \cite{lorin2008clinical} to explain the patient's case. We observed concrete examples of the explanation components being used in the explanations provided by clinicians, such as \textit{`contextual knowledge'} of a patient's condition being used for diagnosis and drug \textit{`recommendations'}. Other examples of  explanation types needed within the system design include \textit{trace-based explanations} in a treatment planning mode, to provide an algorithmic breakdown of the guideline steps that led to a drug recommendation; \textit{`scientific explanations'} in a plan critiquing mode, to provide references to studies that support the drug, as well as \textit{`counterfactual explanations'}, to allow clinicians to add/edit information to view a change in the recommendation; and \textit{`contrastive explanations'} in a differential diagnosis mode, to provide an intuition about which drug is the most recommended for the patient. The results of the user studies demonstrated the need for a diverse set of explanation types and that modeling explanation requires various components to support AI system design. 

An example of our ontology being used to represent the generation process for a \textit{`contrastive explanation'}, while accounting for the \textit{`reasoning mode,'} \textit{`AI Task'} involved, can be viewed in Listing \ref{lst:contrastiveexp}. In the RDF representation of a \textit{`contrastive explanation'} used by a clinician, we depict how our ontology would be useful to guide a system to provide an explanation in real-time to the question, ``Why Drug B over Drug A?''
In our discussion of the Listing \ref{lst:contrastiveexp} hereafter, we include the entity IRIs from the listing in parantheses. Upon identifying what explanation type would best suit this question, our ontology would guide the system to access different forms of \textit{`knowledge'} and invoke the corresponding \textit{`AI tasks'} that are suited to generate \textit{`contrastive explanations'}. In this example, a deductive AI task (:AITaskExample) is summoned and generates a system recommendation (:SystemRecExampleA) that Drug A is insufficient based on contextual knowledge of the patient record (:ContextualKnowledgePatient). In addition, the deductive task is also fed with guideline evidence that Drug B is a preferred drug, which results in the generation of a recommendation (:SystemRecExampleB) in favor of Drug B. Finally, our ontology would help guide a system to populate the components of a \textit{`contrastive explanation'} from \textit{`facts'} that supported the hypothesis, ``Why Drug B?'' and its \textit{`foil'}, ``Why not Drug A?,'' or the facts that ruled out Drug A. We note that the annotation of granular content, such as patient and drug data within these explanations, would require the usage of domain-specific ontologies, whose concepts would need to be inferred into classes supported within our ontology. We defer the granular content annotation effort to future work. 

\lstset{language=Manchester, basicstyle=\ttfamily\fontsize{9}{10}\selectfont, columns=fullflexible, xleftmargin=5mm, framexleftmargin=5mm, numbers=left, numberstyle=\ttfamily\fontsize{9}{10}\selectfont, numbersep=5pt, showstringspaces=false, frame=lines, captionpos=t, caption={Turtle representation of the process a system would undergo to generate a \textit{`contrastive explanation'}, such as the one presented during our cognitive walk-through to address, ``Why drug B over drug A?''}, label=lst:contrastiveexp}
\begin{lstlisting} 
:ContrastiveQuestion 
    a sio:`question';
    rdfs:label ``Why Drug B over Drug A?'' .

:ContrastiveExpInstance 
    a eo:ContrastiveExplanation;
    ep:isBasedOn :SystemRecExampleA, :SystemRecExampleB;
    rdfs:label ``Guidelines recommend Drug B for this patient'';
    :addresses :ContrastiveQuestion .
    
:SystemRecExampleA
    a eo:SystemRecommendation;
    prov:used :ContextualKnowledgePatient;
    rdfs:label ``Drug A is not sufficient for the patient'' . 

:SystemRecExampleB 
    a eo:SystemRecommendation;
    prov:used :GuidelineEvidence;
    rdfs:label ``Drug B is recommended by the guidelines'' . 

:AITaskExample 
    a eo:DeductiveTask;
    sio:`has output' :SystemRecExampleA, :SystemRecExampleB;
    ep:hasSetting [a eo:ReasoningMode; rdfs:label ``Treatment Planning''];
    prov:used :ContextualKnowledgePatient, :GuidelineEvidence;
    rdfs:label ``Deductive task'' .
    
:ContextualKnowledgePatient 
    a eo:ContextualKnowledge, eo:Foil;
    sio:`in relation to' [a sio:`patient'];
    sio:`is input in' :AITaskExample;
    rdfs:label ``patient has hyperglycemia'' .

:GuidelineEvidence 
    a eo:ScientificKnowledge, eo:Fact;
    sio:`is input in' :AITaskExample;
    rdfs:label ``Drug B is the preferred drug'' .
\end{lstlisting}

\section{Evaluation} \label{sec:evaluation}
We evaluate our ontology via a set of competency questions posed from the perspective of a system designer who may need to design a system that includes appropriate explanation support and may hope to use our ontology. These competency questions are designed to aid system designers in their planning of resources to include for generating explanations that are suitable to the expertise level of the end-user, the scenario for which the system is being developed, etc. The resources that a system designer would need to consider could include the \textit{`AI method'} and \textit{`tasks'} capable of generating the set of explanations that best address the user's \textit{`question'} \cite{liao2020questioning}, the reasoning \textit{`modes'} that need to be supported within the system, and the \textit{`knowledge'} sources. These competency questions would be ones that, for example, we, as system designers looking to implement a clinical decision-support system, such as the prototype described in Section \ref{sec:instantiations}, would ask ourselves upon analyzing the requirements of clinicians gathered from a user study. Contrarily, if the specifications were to live in documentation, versus an explanation ontology, it would be cumbersome for a system designer to perform a lookup to regenerate the explanation requirements for every new use case. Through EO we support system designers to fill in what can be thought of as ``slots'' (i.e., instantiate classes) for explanation generation capabilities. 

\begin{table}[hbt!]
\centering
\caption{A catalog of competency questions and candidate answers produced by our ontology. These questions can be generalized to address queries about other explanation types supported within our ontology.}
\label{tab:competencyquestion}
\resizebox{\textwidth}{!}{%
\begin{tabular}{|l|l|}
\hline
\textbf{\large{Competency Question}} &
  \textbf{\large{Answer}} \\ \hline
\begin{tabular}[c]{@{}l@{}}\textbf{Q1. Which AI model (s) is capable of }\\ \textbf{generating this explanation type} \\ 
\textbf{(e.g. trace-based)?} 
\end{tabular} &
  \begin{tabular}[c]{@{}l@{}}Knowledge-based systems,\\ Machine learning model: decision trees\end{tabular} \\ \hline
\begin{tabular}[c]{@{}l@{}}\textbf{Q2. What example questions have been} \\ \textbf{identified for counterfactual explanations?}\end{tabular} &
  \begin{tabular}[c]{@{}l@{}}What other factors about the patient does \\ the system know of?\\ What if the major problem was a fasting \\ plasma  glucose?\end{tabular} \\ \hline
\begin{tabular}[c]{@{}l@{}}\textbf{Q3. What are the components of a} \\\textbf{scientific explanation?}\end{tabular} &
  \begin{tabular}[c]{@{}l@{}}Generated by an AI Task, Based on \\ recommendation, and based on evidence from \\ study or basis from scientific method \end{tabular} \\ \hline
  \begin{tabular}[c]{@{}l@{}}\textbf{Q4. Given the system was performing abductive} \\ \textbf{reasoning and has ranked specific} \\ \textbf{recommendations by comparing different } \\ \textbf{medications, what explanations can be} \\ \textbf{provided for that recommendation?} \end{tabular} &
  Contrastive explanation \\ \hline
\begin{tabular}[c]{@{}l@{}}\textbf{Q5. Which explanation type best suits the } \\ \textbf{user question, ``Which explanation type} \\ \textbf{can expose numerical evidence about} \\ \textbf{ patients on this drug?,'' and how} \\ \textbf{will the system generate the answer?} \end{tabular} &
  \begin{tabular}[c]{@{}l@{}} Explanation type: statistical \\
  System: run `Inductive' AI task with \\ `Clustering' method to generate \\ numerical evidence\end{tabular} \\ \hline
\end{tabular}%
}
\end{table}

In Table \ref{tab:competencyquestion}, we present a list of competency questions along with answers. The first three questions can be addressed before or during system development, when the system designer has gathered user requirements, and the last two questions need to be answered in real-time, based on the system and a user's current set of attributes. During system development and after the completion of a requirements gathering session, if a system designer learns that a certain set of explanation types would be suitable to include, through answers to the first three questions, they can be made aware of the AI methods capable of generating explanations of the type (Q1), the components to build into their system to ensure that the explanations can be generated (Q3), and some of the questions that have been addressed by the particular explanation type (Q2). When a system is running in real-time, an answer to (Q4) would help a system designer decide what pre-canned explanation type already supported by the system would be the best current set of system, and user attributes and an answer to (Q5) would help a system designer decide on whether their system, with its current capabilities, can generate an explanation type that is most suitable to the form of the question being asked. While we address these competency questions, and specifically ask questions Q4 and Q5 in the setting of our clinical requirements gathering session, we expect that these questions can be easily adapted for other settings to be addressed with the aid of EO. In addition to presenting natural-language answers to sample competency questions, we depict a SPARQL query used to address the third question in Listing \ref{lst:sparqlquery}. 

\lstset{language=SPARQL, basicstyle=\ttfamily\fontsize{9}{10}\selectfont, xleftmargin=5mm, framexleftmargin=5mm, numbers=left, stepnumber=1, breaklines=true, breakatwhitespace=false, numbersep=5pt, tabsize=2, frame=lines, captionpos=t, caption={A SPARQL query that retrieves the sufficiency conditions encoded on an explanation type to answer a competency question of the kind, ``What are the components of a scientific explanation?''}, label=lst:sparqlquery}
\begin{lstlisting} 
prefix rdfs:<http://www.w3.org/2000/01/rdf-schema#>
prefix owl:<http://www.w3.org/2002/07/owl#>

select ?class ?restriction 
where {
    ?class (rdfs:subClassOf|owl:equivalentClass) ?restriction .
    ?class rdfs:label ``Scientific Explanation'' .
}
\end{lstlisting}

\begin{table}[hbt!] 
\centering
\caption{Results of the SPARQL query to retrieve the sufficiency conditions defined on the \textit{`scientific explanations'} class. These results depict the flexibility that we allow so that \textit{`scientific explanations'} can either be directly based on \textit{`scientific knowledge'} or on system recommendations that use \textit{`scientific knowledge'}.}
\label{tab:resultstable}
\resizebox{\textwidth}{!}{%
\begin{tabular}{|l|l|l|}
\hline
\textbf{subject}                 & \textbf{restriction}                                                                                                       \\ \hline
Scientific Explanation &  \begin{tabular}[c]{@{}l@{}}(ep:isBasedOn some (eo:`Scientific Knowledge' and ((prov:wasGeneratedBy \\ some `Study')  or (prov:wasAssociatedWith some eo:`Scientific Method'))) and\\  (isBasedOn some eo:`System Recommendation')) or \\
(ep:isBasedOn some (eo:`System Recommendation' and \\ (prov:used some (eo:`Scientific Knowledge' and ((prov:wasGeneratedBy \\ some `Study') or (prov:wasAssociatedWith some \\ eo:`Scientific Method'))))))\end{tabular} \\ \hline
\end{tabular}%
}
\end{table}

\section{Resource Contributions}
\label{sec:contributions}
We contribute the following publicly available artifacts: \textbf{Explanation Ontology} with the \textbf{logical formalizations of the different explanation types} and \textbf{SPARQL queries} to evaluate the competency questions, along with the applicable documentation available on our resource website.
These resources, listed in Table \ref{tab:resourcetable}, are useful for anyone interested in building explanation facilities into their systems. The ontology has been made available as an open-source artifact under the Apache 2.0 license \cite{apache2}
and we maintain an open source Github repository for all our artifacts. We also maintain a persistent URL for our ontology hosted on the PURL service.

\begin{table}[hbt!]
\centering
\caption{Links to resources we have released and refer to in the paper.}
\label{tab:resourcetable}
\resizebox{\textwidth}{!}{%
\begin{tabular}{|l|l|}
\hline
Resource             & Link to resource                                                  \\ \hline
Resource Website   & \url{http://tetherless-world.github.io/explanation-ontology}   \\ \hline
EO PURL URL        & \url{https://purl.org/heals/eo}                                \\ \hline
Github Repository  & \url{https://github.com/tetherless-world/explanation-ontology} \\ \hline
\end{tabular}%
}
\end{table}

\section{Discussion and Future Work} \label{sec:discussion}
To address the gap in a semantic representation that can be used to support the generation of different explanation types, we designed 
an OWL ontology, an explanation ontology, that can be used by system designers to incorporate different explanation types into their AI-enabled systems. 
We leverage and maintain compatibility with an existing explanation patterns ontology \cite{tiddi2015ontology} and we expand on it to include representational primitives needed for modeling explanation types for system designers.
We also leveraged the widely-used general-purpose SIO ontology, and introduce the classes and properties necessary for a system to generate a variety of user-centric explanations. During our modeling, we make certain decisions that are inspired by our literature-review and knowledge of the usage of explanation types in clinical practice, to include classes (e.g., \textit{`system recommendation,'} \textit{`knowledge,'} \textit{`user'}) that we deem as central to generating explanations. We include other classes that would indirectly be needed to generate explanations and reason about them, hence, capturing the process involved in explanation generation. However, through our ontology we do not generate natural language explanations, and rather provide support to fill in ``slots'' that will be included in them. 

Our explanation ontology is comprehensive and flexible as it was designed from requirements gathered from a relatively extensive literature review along with a requirements gathering session in a clinical domain. In this paper, we have described how the ontology can be used to represent literature-derived explanation types and then how those explanation types address the questions posed by clinicians. The ontology is also designed to be extensible, as with all ontologies, representational needs may arise as applications arise and evolve. Our competency questions provide guidance to system designers as they make their plans for providing explanations within their decision support systems. 

In the future, we plan to build a middleware framework (such as the service we alluded to in Section \ref{sec:introduction}) that would interact with the system designer, take a user's \textit{`question'} as input, and apply learning techniques on a combination of the user's \textit{`question'} and the inputs available to the AI system, which could include the user's \textit{`situation'} and context, and 
the system's \textit{`reasoning mode'} to decide on the most suitable \textit{`explanation'} type. 
Upon identifying the appropriate \textit{`explanation'} type, the framework would leverage the sufficiency conditions encoded in our ontology to gather different forms of \textit{`knowledge'} to generate the suitable explanation type and summon the AI \textit{`tasks'} and \textit{`methods'} that are best suited to generating the explanation. 
Such a framework would then be capable of working in tandem with hybrid AI reasoners to generate hybrid explanations \cite{chari2020directionschapter,mittelstadt2019explaining} that serve the users' requirements. 

We have represented the explainability components that we deem necessary to generate explanations with a user focus. However, there are other aspects of user input that 
may be harder to capture. Wang et al. \cite{wang2019designing} have shown that there is a parallel between one of these user aspects, a user's reasoning strategies, and the reasoning types that an \textit{‘AI Task'} uses to support the generation of explanations to address these situations. We are investigating how to include classes, such as a user's reasoning strategies, that are harder to capture/infer from a system perspective and would be hard to operationalize in a system's model. The user-centric focus of AI has been emphasized recently by Liao et al. in their question bank of various user questions around explainability that suggests ``user needs should be understood, prioritized and addressed'' \cite{liao2020questioning}. As we start to build more user attributes into our ontology, we believe that our model will evolve to support more human-in-the-loop AI systems. We are using EO as a foundation for generating different explanation types in designing a clinical decision support system, and we will publish updates to our ontology as we make edits to support new terms.

\section{Conclusion} \label{sec:conclusion}
We have built an ontology aimed at modeling explanation primitives that can support user-centered AI system design.  Our requirements came from a breadth of literature and requirements gathered by prospective and actual users of clinical decision support systems. We encode sufficiency conditions encapsulating components necessary to compose hybrid explanation types that address different goals and expose different forms of knowledge in our ontology. Through a carefully crafted set of competency questions, we have exposed and evaluated the coverage of our ontology in helping system designers make decisions about explanations to include in their systems. We believe our ontology can be a valuable resource for system designers as they plan for what kinds of explanations their systems will need to support. We are continuing to work towards supporting the generation of different explanation types and designing a service that would use our explanation ontology as a base to generate explanation types that address users' questions. 

\section*{Acknowledgements}
This work is done as part of the HEALS project, and is partially supported by IBM Research AI through the AI Horizons Network. We thank our colleagues from IBM Research, including Ching-Hua Chen, and from RPI, Sabbir M. Rashid, Henrique Santos, James P. McCusker and Rebecca Cowan, who provided insight and expertise that greatly assisted the research.

\bibliographystyle{splncs04}
\bibliography{references}

\end{document}